\documentclass[]{ieeeaccess}

\usepackage{cite}
\usepackage{amsmath,amsfonts,amssymb,amsthm}
\usepackage[pdftex]{graphicx}
\usepackage{amsbsy,verbatim}
\usepackage{times}
\usepackage[english]{babel}
\usepackage{stfloats}
\usepackage{dsfont}
\usepackage{url}
\usepackage[format=default,labelsep=colon,labelfont=footnotesize,font=footnotesize]{caption}
\usepackage[caption=false]{subfig}
\usepackage{booktabs}
\usepackage{flushend}
\usepackage{algorithm}
\usepackage[noend]{algpseudocode}

\newcommand{\wwmpg}{0.485\linewidth}

\DeclareMathOperator*{\argmin}{arg\,min}

\makeatletter
\newcommand\remembertext[2]{
  \immediate\write\@auxout{\unexpanded{\global\long\@namedef{mytext@#1}{#2}}}%
  {\color{blue}{#2}}%
}
\newcommand\recalltext[1]{%
  \ifcsname mytext@#1\endcsname
    \@nameuse{mytext@#1}%
  \else
    ``???''
  \fi
}
\makeatother

\history{Received xxxxxx xx, xxxx, accepted xxxxxx xx, xxxx. Date of publication xxxx xx, xxxx, date of current version xxxx xx, xxxx.}
\doi{10.1109/ACCESS.201x.xxxxxxx}

\title{A Novel Hyperparameter-free Approach to Decision Tree Construction that Avoids Overfitting by Design}

\author{\uppercase{Rafael~Garc\'{\i}a~Leiva}, \uppercase{Antonio~Fernandez~Anta}~\IEEEmembership{Senior Member,~IEEE}, \uppercase{Vincenzo~Mancuso}~\IEEEmembership{Member,~IEEE}, \uppercase{Paolo~Casari}~\IEEEmembership{Senior Member,~IEEE}\vspace{1.5mm}}
\address[]{IMDEA Networks Institute, 28918 Madrid, Spain}
\corresp{Corresponding author: Rafael~Garc\'{\i}a~Leiva (e-mail: rafael.garcia@imdea.org).}

\date{}

\begin{document}

\vol{X}
\year{20XX}

\begin{abstract}
Decision trees are an extremely popular machine learning technique. Unfortunately, overfitting in decision trees still remains an open issue that sometimes prevents achieving good performance. In this work, we present a novel approach for the construction of decision trees that avoids the overfitting by design, without losing accuracy. A distinctive feature of our algorithm is that it requires neither the optimization of any hyperparameters, nor the use of regularization techniques, thus significantly reducing the decision tree training time. Moreover, our algorithm produces much smaller and shallower trees than traditional algorithms, facilitating the interpretability of the resulting models.\vspace{5mm}
\end{abstract}

\begin{IEEEkeywords}
\noindent Decision Trees, Regularization, Interpretability, Kolmogorov Complexity\vspace{5mm}
\end{IEEEkeywords}

\maketitle

\thispagestyle{empty}

%
%
\section{Introduction}
\label{sec:introduction}

Decision trees are a highly popular machine learning technique that has been successfully applied to solve a number of practical problems in areas such as natural language processing, information extraction, data mining, and pattern recognition~\cite{lior2014data}. Decision trees are also the foundation of more advanced machine learning methods, such as random forests or boosted trees~\cite{friedman2001elements}.

One of the most challenging tasks for decision tree-building algorithms is to decide when to stop the tree growing process. Most of the tree building methods produce very complex models that \emph{overfit} training data. Overfitted trees not only have poor predictive capabilities on newpreviously unseen data, but also can be exceedingly difficult to interpret, which is a key barrier against the adoption of these models in practical applications~\cite{wu2017beyond}. 
A common approach to avoid overfitting in decision trees is \emph{early stopping}, which forces the construction algorithm to stop before the tree becomes too complex. Popular stopping criteria include limiting the maximum depth of the tree, requiring a minimum number of sample points at leaf nodes, or computing the accuracy gain yielded by new nodes~\cite{lior2014data}. However, these heuristics require the optimization of (possibly multiple) hyperparameters, which makes the training process computationally expensive. 

An alternative approach to avoid overfitting in decision trees is to apply a post-processing \emph{pruning} algorithm that removes those branches that contribute the least to the accuracy of the final model~\cite{bohanec1994trading}. Cost-complexity~\cite{friedman2001elements} is a common pruning technique, based on applying a cost function that takes into account how well the tree fits the training data and a regularization factor based on how complex the tree is. If we denote by $E(T)$ the fitness of a tree $T$, and by $L(T)$ its complexity, the cost function $C(T)$ could be given by
\begin{equation}
C(T) = E(T) + \lambda L(T),
\end{equation}
where $\lambda$ is a configurable parameter that controls the trade-off between fitness and complexity. In practice, the accuracy of the model is typically used as a fitness metric, and the number of leaf nodes in the tree as a complexity metric. Although, on average, cost complexity pruning produces better results than limiting the growth of the tree via early stopping~\cite{ripley1996pattern}, this technique requires the optimization of the hyperparameter $\lambda$.

In this paper we propose a decision tree construction algorithm with early stopping properties, that does not require hyperparameter tuning or pruning in order to get good accuracy and trees of small size.

\subsection{Related Work}
\label{sec:relatedWork}
 
Earlier algorithms for the construction of decision trees were based on a exhaustive search of the space of possible trees (see ~\cite{loh2014fifty} for a historical review of the discipline). However, due to the computational complexity of those algorithms, alternative approaches were investigated. Most of the modern algorithms used in practice are based on greedy search techniques (see for example ~\cite{cormen2009introduction}), that significantly reduce the training time, at the cost of producing sub-optimal trees.

Among the algorithms to build decision trees, the Iterative Dichotomiser~3 algorithm, ID3~\cite{quinlan1986induction}, is based on an information gain splitting criteria, which stops when all the training samples belong to a single target value, or when no possible split has a positive information gain. Classification And Regression Trees, CART~\cite{breiman1984classification}, builds binary trees based on a growing splitting criteria and applying a cost-complexity pruning. CHi-squared Automatic Interaction Detector, CHAID~\cite{kass1980exploratory}, relies on an adjusted significance tests to find the least significant values with respect to the target attribute, and the process is repeated until no significant pairs are found. C4.5~\cite{quinlan1993c4} is an evolution of ID3 that uses gain ratio as a splitting criteria, and ceases to grow the tree when the number of instances to split is below a given threshold. It includes an accuracy based post-pruning. The Quick Unbiased Efficient Statistical Tree algorithm, QUEST~\cite{loh1997split}, uses quadratic discriminant analysis for splits, the stopping is based on statistical tests, and the resulting trees are pruned based on cross-validation. Multivariate Adaptive Regression Splines, MARS~\cite{friedman1991multivariate}, is based on a two-phase (forward and backward pass) algorithm that uses regression functions approximated using linear splines and their tensor products. PUBLIC~\cite{rastogi1998public} is based on minimizing the sum of the length of the shortest computer program that can print the data given as input to the tree, and the length of the shortest program that can print the tree. Recent trends in tree building algorithms include deep trees~\cite{ignatov2017decision} of hundred of levels, and oblique trees~\cite{carreira2018alternating} with more advanced if-else conditions. A detailed comparison of decision trees algorithms can be found in~\cite{lim2000comparison}, and an up-to-date review including other classification methods in~\cite{zhang2017up}.

\subsection{Contributions}

Unlike in the literature surveyed above, the main contribution of this work is a novel algorithm to construct decision trees that, by design, do not overfit the training data. The algorithm is based on a novel cost function that is used to decide when to stop the building process, rather than to indicate how to prune the generated tree. Our algorithm does not depend on the optimization of hyperparameters, thus considerably reducing the training time. This means that, with our algorithm, we can process a very significant amount of data with reduced complexity, e.g., by using the full dataset for training, or by enacting a training / validation split if a better estimate of the real accuracy of the resulting model is sought.

The models generated with our algorithm are significantly smaller in terms of number of nodes, and more shallow than the models generated by other decision tree building algorithms; at the same time, the obtained models present similar, and sometimes better accuracy. Shallower models allow us to make faster predictions when they are used as part of large ensembles of trees, and largely improve interpretability, e.g., when examined by domain experts. Additional advantages of our new algorithm include lower sensitivity to the presence of errors in the training dataset and to the non-linearity of the hyper-boundaries between different classes.


%
%
\section{Notation and Background}

\subsection{Notation}

Let $\mathcal{X}$ be a dataset composed of $n$ training input vectors $x_i \in \mathbb{R}^m$, were $1 \leq i \leq n$, and let $y \in \mathcal{G}^n$ be a category vector, where $\mathcal{G}$ is the set of classification labels ($\mathcal{G} = \{0, 1, \ldots, G\}$). The problem at hand is to find the best model $f$, from a given family of models $\mathcal{F}$, such that $f(x_i) = y_i$ for \emph{many} $x_i \in \mathcal{X}$. That is, we are interested in solving a \emph{supervised classification problem}. We are interested in the capability of the model $f$ to generalize to previously unseen data, that is, to correctly classify input vectors extracted from a universe $\mathcal{U}$ of input vectors, not necessarily included in the training dataset $\mathcal{X}$. For this reason,
the best solution is usually not the overfitted model $f$ that guarantees $f(x_i) = y_i$ for \emph{all} $x_i \in \mathcal{X}$.

\subsection{Kolmogorov Complexity}

The novel cost function introduced in this paper to construct decision trees is based on the concept of Kolmogorov complexity~\cite{solomonoff1964formal,kolmogorov1965three,chaitin1969simplicity}, also known as Algorithmic Information Theory. The application of Kolmogorov complexity to the search of optimal statistical models is implemented by the Minimum Description Length (MDL) principle~\cite{grunwald2007minimum} and the Minimum Message Length (MML)~\cite{wallace2005statistical}. In particular, minimum length techniques have been applied to the problem of inferring decision trees in~\cite{quinlan1989inferring}, later on clarified and extended in~\cite{wallace1993coding}, as well as in~\cite{mehta1995mdl} as a technique for pruning; in~\cite{rastogi1998public} the PUBLIC algorithm based on the MML is described. Although the underlining concepts behind our own cost function are the same (namely, that learning is equivalent to the capability to compress), our approach is very different from the ones described above.

According to the Kolmogorov complexity, the amount of information of an object, encoded as a finite string, is given by the length of the shortest computer program that is able to print (output) that string. Kolmogorov complexity does not require to know the set of valid strings in advance or to make assumptions about their probability distribution. Therefore, it provides a universal definition of the amount of information contained in an object. However, the Kolmogorov complexity is a non-computable quantity~\cite{li2008introduction}, hence it must be approximated in practice.

The \emph{Kolmogorov complexity} of a string $s$ composed by symbols from a fixed alphabet $\Sigma$, denoted by $K(s)$, is defined 
as\footnote{$|x|$ denotes the number of symbols, or length, of a string $x$.}
\begin{equation}
K(s)=\min_{p,v}\big\{ |p| + |v|\,:\, U(p,v)=s\big\},
\end{equation}
where $U$ is a universal computer, $p$ is a program in a prefix-free language interpreted by $U$,%
\footnote{A programming language is prefix-free if no program can be a prefix of another program.}
and $v$ is the input to the program. 
It can be shown that the Kolmogorov complexity $K(s)$ of a string $s$ does not depend on the programming language used~\cite{li2008introduction}, i.e., any reasonable and sufficiently powerful computer language provides the same description length, up to a fixed additive constant that depends on the selected language, but not on the string itself.

Sometimes, the description of a string can be greatly reduced if we assume the knowledge of another string. The \emph{conditional Kolmogorov complexity} of a string $s$ given the string $s'$, denoted by $K(s | s')$, is defined as\footnote{$\langle v, s' \rangle$ denotes the concatenation of the strings $v$ and $s'$}
\begin{equation}
K(s|s')=\min_{p,v}\big\{ |p| + |v|\,:\, U(p,\langle v, s' \rangle)=s\big\}.
\end{equation}

\section{Decision Tree Construction Algorithm}

\subsection{Key Idea}

In this paper, we provide an algorithm to construct decision trees that, by design, does not overfit training data, and that has no hyperparameters to be optimized. To achieve this, the algorithm must automatically understand when growing the decision tree adds needless complexity, and must measure such complexity in a way that is commensurate to some prediction quality aspect, e.g., inaccuracy. We argue that a natural way to achieve the above objectives is to define both the inaccuracy and the complexity (conveyed by the so-called \emph{surfeit}, introduced later on) using the concept of Kolmogorov complexity~\cite{solomonoff1964formal}.

The key insight behind the application of Kolmogorov complexity to machine learning is that the more patterns we can find in a dataset, the more we have learned about the data. This means both identifying the dataset features that best explain the outcome, and to devise a model that provides this explanation without exceeding complexity. Moreover, data compression is about finding and exploiting patterns and regularities in the data. A practical implementation that puts together the key features of Kolmogorov complexity and data compression follows the lines of the Minimum Description Length principle~\cite{grunwald2007minimum}, that proposes to minimize the length of the model plus the length of the data given the model, namely $L(M) + L(D \! \mid \! M)$, where $M$ denotes a model and $L(\cdot)$ returns the length of its argument. Unfortunately, this approach tends to favor simple models, and is not widely used in practice due to this limitation. 

In this work, we propose to replace the minimization of $L(M) + L(D \! \mid \! M)$ by a multiobjective optimization problem~\cite{miettinen2012nonlinear} that reconciles two conflicting goals: to minimize the complexity of the model, and to minimize the inaccuracy of the model's predictions. Moreover, instead of working directly with the length of models $L(M)$ we propose to work with their surfeit, which evaluates the unnecessary complexity of a model used to represent a given dataset. The surfeit can be computed as the length of the model minus the length of the minimum computer program that can print the dataset, i.e., the Kolmogorov complexity of the data. Namely, this is $L(M) - K(D)$. Using the surfeit avoids the rejection of correct but complex models. Our approach is validated in practice by means of its application to a family of decision tree models.

\subsection{Overview of the Algorithm}

We describe now the algorithm (called Minimum Surfeit and Inaccuracy, or MSI for short) to build a decision tree given a training dataset $\mathcal{X}$ with the help of the pseudocode reported in Algorithm~\ref{algorithm:decision_tree}. The algorithm is based on a breadth-first tree traversal, see for example~\cite{cormen2009introduction}. The algorithm requires a function called \textsc{bestSplit()}, that returns the best way to split a given subset of the training data into two subsets, and a second function called \textsc{costFunc()}, that provides a quantitative evaluation of the cost of a tree in terms of complexity and accuracy. The details of these two functions are given in Sections~\ref{sub:splitting_criteria} and~\ref{sub:cost_function}, respectively. A third function \textsc{Forecast()} computes the most likely class of a given subset of the training data. Algorithm~\ref{algorithm:decision_tree} is based on two nested loops: the external \textbf{while} loop keeps a set $\mathit{Candidates}$ of the candidate tree nodes (leaves) to grow, whereas the internal \textbf{for each} loop finds the best such node from which the tree should be grown further. The latter operation requires to check all possible options and select the one that minimizes the cost of the resulting tree. The exit point in the algorithm is at the end of the \textbf{while} loop, where the current tree $T$ is returned if there are no more candidate nodes to further grow the tree, or the nodes in the set $\mathit{Candidates}$ generate trees that do not reduce the cost.
The nodes of the tree are represented with the \textit{TreeNode} data structure composed of the elements (1) \textit{LChild}, that points to the left child node in the tree, (2) \textit{RChild}, that points to the right child node, (3) \textit{Data}, that contains the subset of training data of the node, (4) \textit{Split}, which is a pair composed by the feature and a threshold used for the data splitting, and (5) \textit{Class}, which is the most likely class for this node (based on its \textit{Data}). A tree node could be either an internal node having a split criterion and two children, or a leaf node with a predicted class and no child.

\begin{algorithm*}[ht!]
\caption{Minimum Surfeit and Inaccuracy (MSI)}
\label{algorithm:decision_tree}
\begin{algorithmic}[1]

\Procedure{buildTree}{$data$}

    \State Create TreeNode $r$; $r.\mathit{LChild}, r.\mathit{RChild}, r.Split \gets \textbf{none}$; $r.Data \gets data$; $r.\mathit{Class} \gets \Call{forecast}{data}$

    \State Create Tree $T$; $T.Root \gets r$; $bestCost \gets \Call{costFunc}{T}$

    \State $\mathit{Candidates} \gets \{r\}$

    \While {$\mathit{Candidates} \neq \emptyset$}
    

        \State $\mathit{bestSol} \gets \textbf{none}$

        \For{\textbf{each} $\ell \in \mathit{Candidates}$}

            \State $\Theta \gets \Call{bestSplit}{\ell.Data}$
            
            \If {$\Theta = \emptyset$} $\mathit{Candidates} \gets \mathit{Candidates} \setminus \{\ell\}$ \Comment{Discard $\ell$ as candidate to grow the tree.}
            \Else
            
                \State Create TreeNode $\ell L, \ell R$ \Comment{Tree nodes $\ell L, \ell R$ will be the children of $\ell$.}
                
                \State $\ell L.\mathit{LChild}, \ell L.\mathit{RChild}, \ell L.Split, \ell R.\mathit{LChild}, \ell R.\mathit{RChild}, \ell R.Split \gets \textbf{none}$
                
                \State $(dataL, dataR) \gets \Call{split}{\Theta, \ell.Data}$; $\ell L.\mathit{Data} \gets dataL$; $\ell R.\mathit{Data} \gets dataR$
                
                \State $\ell L.\mathit{Class} \gets \Call{forecast}{dataL}$; $\ell R.\mathit{Class} \gets \Call{forecast}{dataR}$

                \State $\langle \ell.\mathit{LChild}, \ell.\mathit{RChild}, \ell.Split \rangle \gets \langle \ell L, \ell R, \Theta \rangle$ \Comment{Grow the tree at node $\ell$.}
 
                \State $C \gets \Call{costFunct}{T}$; \Comment{ $C$ is the cost of the tree if grown at $\ell$.}
                
                \State $\langle \ell.\mathit{LChild}, \ell.\mathit{RChild}, \ell.Split \rangle \gets \langle \textbf{none}, \textbf{none}, \textbf{none} \rangle$  \Comment{Leave the tree as it was.}
                
                \If {$C < bestCost$} $bestCost \gets C$; $\mathit{bestSol} \gets (\ell, \Theta, \ell L, \ell R)$

                \EndIf

            \EndIf
            
        \EndFor
            
        \If {$\mathit{bestSol} \neq \textbf{none}$}

            \State $(\ell, \Theta, \ell L, \ell R) \gets \mathit{bestSol}$
            
            \State $\langle \ell.\mathit{LChild}, \ell.\mathit{RChild}, \ell.Split \rangle \gets \langle \ell L, \ell R, \Theta \rangle$

            \State $\mathit{Candidates} \gets \mathit{Candidates} \setminus \{ \ell \} \cup \{\ell L, \ell R\}$
            
        \Else ~\Return $T$ \Comment{No candidate reduces the cost of the tree.}
        
        \EndIf
        
    \EndWhile
    
    \State \Return $T$ \Comment{No more candidates.}

\EndProcedure
\end{algorithmic}
\end{algorithm*}

The main difference between our algorithm and other decision tree algorithms is in the \textbf{for each} loop. In traditional algorithms, the order in which the branches are evaluated is irrelevant. 
However, since our algorithm could stop the growing process at any point, at each iteration we explicitly select the best candidate node to grow the tree.

\subsection{Splitting Criterion}
\label{sub:splitting_criteria}

A decision tree is an algorithm that recursively partitions the training vectors of $\mathcal{X}$ in such a way that the same values $y_i$ are grouped together. Given a subset $Q \subseteq \mathcal{X}$ we have to find the optimal split for $Q$. A split is a pair $\theta = (j,w)$ were $1 \leq j \leq m$ is an index and $w \in \mathbb{R}$ a threshold. A split partitions the set $Q$ into two disjoint subsets $Q_l = \{x_i \in Q : x_{ij} \leq w\}$, and $Q_r = Q \backslash Q_l$. We use  the \emph{minimal weighted entropy} as the splitting criterion.%
\footnote{Other metrics, like Gini impurity or information gain could be used for the same purpose.} 
The weighted entropy of a split, denoted by $\tilde{H}$, is defined as:
\begin{equation}
\tilde{H}(Q, \theta) = \frac{d(Q_l)}{d(Q)} H(Q_l) + \frac{d(Q_r)}{d(Q)} H(Q_r)
\label{equation:weighted_entropy}
\end{equation}
where $d(S)$ is the number of elements (or diameter) of set $S$, and $H(S)$ is the entropy of $S$, i.e., $H(S) = - \sum_{x \in S} p(x) \log p(x)$, being $p(x)$ the probability of getting $x$ if we select a random element from $S$. 
The function \textsc{bestSplit}$(Q)$ returns the best split $\theta^\star$ of a given subset of data $Q$, defined as $\theta^\star = \argmin_\theta \tilde{H}(Q,\theta)$. If $\theta^\star$ does not split $Q$, \textsc{bestSplit}$(Q)$ returns $\emptyset$.

\subsection{Cost Function}
\label{sub:cost_function}

For every possible branch to grow, we have to compute how good the resulting tree would be if we add the new nodes, compared to the same tree without them. In this section we introduce a novel cost function to evaluate and compare candidate trees. A problem with traditional cost functions for tree evaluation is that they are based on incommensurable quantities, such as the accuracy of the model and the number of nodes. Our goal here is to introduce two new metrics that are conceptually equivalent to the traditional ones while still being commensurable, i.e., based on the same units. We have also designed them to have the same scale, in order to enable a direct comparison. These two metrics will convey how badly the tree fits the data (inaccuracy), and how unnecessarily complex the model is (what we call \textit{surfeit}).

Our measure of the inaccuracy of a model $M$ for a dataset $\mathcal{X}$ will be the length of the shortest computer program that, given as input the model $M$, is able to print the dataset $\mathcal{X}$. We seek to measure how difficult (in terms of the size of a program, not its running time) it is to fix the errors introduced by the model. Intuitively, in this sense, it is more difficult to fix a model that makes one hundred different mistakes than a model that makes one hundred times the same mistake. Formally, the \emph{inaccuracy} of a dataset $\mathcal{X}$ and a candidate model $M$ is
\begin{equation}
\mathcal{I}(\mathcal{X}, M) = \frac{ K(\mathcal{X} | M) }{ K(\mathcal{X}) },
\end{equation}
where $K(\mathcal{X} | M)$ is the conditional Kolmogorov complexity of the dataset $\mathcal{X}$ given the model $M$. The normalization factor $K(\mathcal{X})$, the Kolmogorov complexity of the dataset $\mathcal{X}$, is introduced to guarantee that both the inaccuracy and the surfeit have the same scale. We recall that Kolmogorov complexity is a non-computable quantity. Therefore, we approximate $\mathcal{I}(\mathcal{X}, M)$ by the ratio $|{\it Comp}(E)| / |{\it Comp}(\mathcal{X})|$, where $|{\it Comp}(E)|$ is the length of the compressed version of the subset $E \subset \mathcal{X}$ composed by those points that have been misclassified by the tree, and $|{\it Comp}(\mathcal{X})|$ is the length of the compressed version of the full dataset.

To assess the model's complexity, we seek to measure  the amount of surfeit introduced by our current model with respect to the shortest possible model. Formally, the \emph{surfeit} of a dataset $\mathcal{X}$ and a candidate model $M$ is
\begin{equation}
\mathcal{S}(\mathcal{X}, M) = 1 - \frac{K(\mathcal{X})}{|M|}.
\end{equation}
The length of the shortest possible model for the dataset $\mathcal{X}$ is given by its Kolmogorov complexity $K(\mathcal{X})$. The normalization factor is given by the length of the current model being evaluated, $|M|$. Formally, exceedingly short models $M$ such that $|M| < K(\mathcal{X})$, are not considered candidate models. Again, because $K(\mathcal{X})$ is a non-computable quantity, in practice we approximate the surfeit of a model by its redundancy, that is, by computing $1 - |Comp(M)| / |M|$, where $|Comp(M)|$ is the length of a compressed version of a string describing the tree $M$. We remark that the shortest possible model $M$ for a dataset $\mathcal{X}$ must be incompressible, but an incompressible model for $\mathcal{X}$ is not necessarily the shortest possible one. That is, it might happen that a model is not redundant but still presents some surfeit.

Both quantities, inaccuracy and surfeit, have to be combined into a single value by applying a function
\begin{equation}
    \mathcal{N}(\mathcal{X}, M) = g(\mathcal{I}(\mathcal{X}, M), \mathcal{S}(\mathcal{X}, M)) \, .
    \label{eq:n}
\end{equation}
Since both the inaccuracy and the surfeit are relative quantities, a natural way to combine them is via their harmonic mean. Other candidate functions (arithmetic mean, geometric mean, Euclidean distance, product, and addition) will be evaluated in the following, in order to confirm that the selected function provides the best result among these.

\section{Practical Implementation} \label{sec:practical_impl}

The abstract concept of Kolmogov complexity is usually approximated in practice with compression algorithms~\cite{li2008introduction}. In this work we have tested three different algorithms: the Lempel-Ziv-Markov chain algorithm, LZMA~\cite{ziv1977universal} that uses dictionaries for compression, zlib~\cite{deutsch1996deflate} that combines dictionaries with Huffman encodings, and \texttt{bz2} a compressor based on the Burrows-Wheeler transform~\cite{burrows1994block}. All compressors have been configured to use the maximum compression level allowed, in order to avoid problems due to small window size buffers~\cite{cebrian2005common}.

For the representation of a tree as a string we use the following template:

\begin{verbatim}
def tree{[attrs]}:
    if [attr] <= [thresh]:
        return [label] || [subtree]
    else:
        return [label] || [subtree]
\end{verbatim}

Where \texttt{[attrs]} is the list of attributes used, and only those used in the model,%
\footnote{If the dataset contains many attributes, listing all of them when dealing with very short models would make the length of the model's header greater than the length of the body.} 
\texttt{[attr]} is a single attribute represented by the letter \texttt{X} followed by a number (e.g. \texttt{X1}), \texttt{[thresh]} is the threshold used for the split, \texttt{[label]} is one of the valid labels from the set $\mathcal{G}$, and \texttt{|| [subtree]} means that the \texttt{return} statement can be replaced by another level of \texttt{[if - else]} conditions. We could have used a much shorter description of trees by replacing word tokens with symbols, e.g., via the ternary conditional operators \texttt{?} and \texttt{:} used in modern programming languages, or by dropping the \texttt{return} statement. This would produce shorter trees, but the complexity of the models would remain the same, up to an additive constant that does not depend on the models themselves. 
Since, e.g., the harmonic mean compares relative values instead of absolute ones, this additive constant can be safely ignored.

%
%
\section{Results}
\label{sec:results}

\begin{figure}[t!]
\centering
\includegraphics[trim = {0 0 0 7mm}, clip,width=\columnwidth]{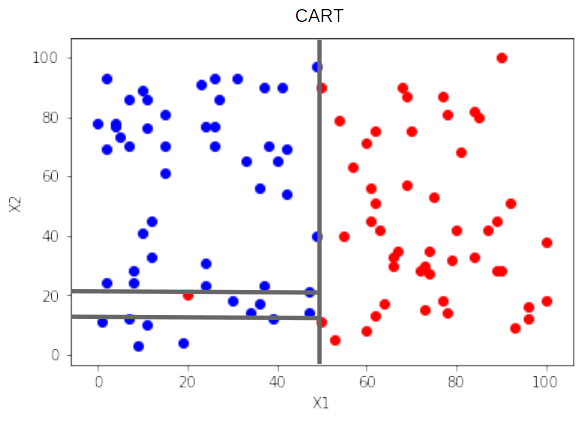}
\caption{Synthetic dataset with splits computed by CART.}
\label{figure:data_error_cart}
\end{figure}
\begin{figure}[t!]
\centering
\includegraphics[width=0.2\textwidth]{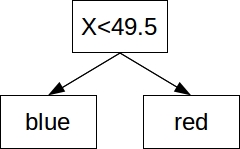}
\caption{Decision tree obtained by our algorithm.}
\label{figure:data_error_nes}
\end{figure}

\begin{figure*}[t]
    \centering
    \begin{minipage}[t]{\wwmpg}
        \centering
        \includegraphics[width=\columnwidth]{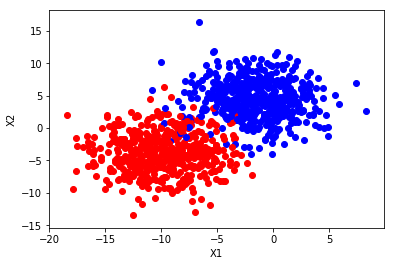}
        \caption{Example of isotropic Gaussian blobs (std. dev. = 3).}
        \label{figure:blobs}
    \end{minipage}
    \hspace{2mm}
    \begin{minipage}[t]{\wwmpg}
        \centering
        \includegraphics[trim={0 0 0 7mm}, clip, width=\columnwidth]{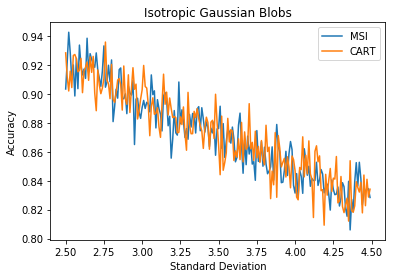}
        \caption{Accuracy for isotropic Gaussian blobs.}
        \label{figure:accuracy_blobs}
    \end{minipage}
\end{figure*}
\begin{figure*}[t]
    \centering
    \begin{minipage}[t]{\wwmpg}
        \centering
        \includegraphics[trim={0 0 0 7mm}, clip, width=\columnwidth]{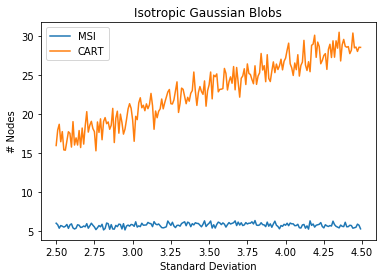}
        \caption{Number of tree nodes for isotropic Gaussian blobs.}
        \label{figure:length_nodes}
    \end{minipage}
    \hspace{2mm}
    \begin{minipage}[t]{\wwmpg}
        \centering
        \includegraphics[trim={0 0 0 7mm}, clip, width=\columnwidth]{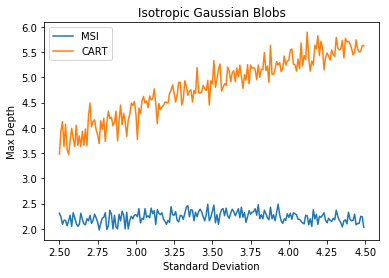}
        \caption{Maximum tree depth for isotropic Gaussian blobs.}
        \label{figure:blob_max_depth}
    \end{minipage}
\end{figure*}

In this section we evaluate our new algorithm, and compare its performance against the well-known algorithms CART, C4.5, CHAID, and PUBLIC. In order to measure prediction performance of the tree classifiers, we used the mean accuracy, defined as one minus the mean error. Other performance parameters taken into account include the total size of the tree, and the maximum depth.

Figure~\ref{figure:data_error_cart} shows a synthetic dataset consisting of 100 random points lying on a two-dimensional plane, where all the points with an $X1$ attribute smaller than $50$ are colored blue, and the rest as red. We artificially introduce a red point, simulating a measurement error, in the blue area. The gray lines correspond to the classification performed by the CART algorithm, as implemented by the scikit-learn toolkit ~\cite{pedregosa2011scikit}. CART will not stop until all the points have been correctly classified, so we require to have at least five points in order to make a node split, a de facto standard suggested, for example, in Section~9.2.2 of~\cite{friedman2001elements} (for a more theoretical justification of this minimum value see the expected count condition at~\cite{starnes2010practice}).

The tree obtained by applying our algorithm to Figure~\ref{figure:data_error_cart} can be seen in Figure~\ref{figure:data_error_nes}. Our algorithm does not try to model the error point, since the gain due to an increment in the accuracy does not compensate the redundancy introduced in the model. Recall that our algorithm stops growing the tree when the total cost function of the tree, based on the measures of inaccuracy and surfeit, does not decrease when adding new nodes to the tree. Our algorithm presents a lower sensitivity to the errors found in datasets, at least if the number of errors is small compared to the number of valid points.

A second experiment with synthetic datasets is depicted in Figure~\ref{figure:blobs}. There, we create two isotropic Gaussian blobs that partially overlap. We start with a standard deviation of $2.5$ for each cluster, so they are easy to separate, and we increase the standard deviation in increments of $0.01$, until we reach $4.5$, which causes significant overlaps. For each value of the standard deviation, we run the experiment $100$ times and we compute the average accuracy for the two algorithms using different datasets for training and testing. For this experiment, the hyperparameter ``minimum number of samples per leaf node'' of the CART algorithm has been optimized in order to achieve the maximum accuracy (in this case, the best value was obtained with a minimum size of 26 samples). The results of this experiment are shown in Figure~\ref{figure:accuracy_blobs}. 
On average, our algorithm provides the same accuracy (0.872 on average) as the optimized version of the CART algorithm.



 For each iteration of the experiment, we have also computed the average number of nodes, including internal and leaf nodes, required by the models to properly classify the clouds in the dataset. The results of this measurement are show in Figure~\ref{figure:length_nodes}. Our algorithm requires an average of $5.7$ nodes compared to $23$ nodes for the CART algorithm. Moreover, our algorithm is more stable than CART, in the sense that it produces models of similar complexity when it gets similar input datasets (a standard deviation of $0.3$ compared to $3.9$ for CART).


In Figure~\ref{figure:blob_max_depth} we show the maximum depth of the tree, defined as the longest path from the root of the tree to any of its leaves. The maximum depth of the tree is a good measure of the maximum time it will require for the model to provide a classification. Our algorithm has an average depth of $2.2$ nodes, whereas the average depth yielded by the CART algorithm is $4.8$ nodes. We emphasize that the CART algorithm requires to optimize a configuration hyperparameter in order to obtain these optimal results, whereas the algorithm we propose, by design, does not require this optimization at all.

\begin{figure*}[t]
    \centering
    \begin{minipage}[t]{\wwmpg}
        \centering
        \includegraphics[trim={0 0 0 7mm}, clip, width=\columnwidth]{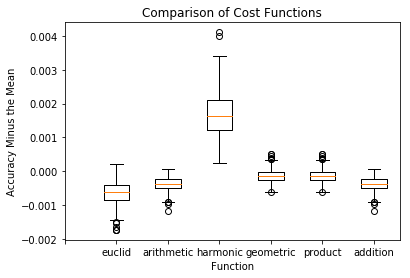}
        \caption{Evaluation of cost functions.}
        \label{figure:metrics}
    \end{minipage}
    \hspace{2mm}
    \begin{minipage}[t]{\wwmpg}
        \centering
        \includegraphics[trim={0 0 0 7mm}, clip, width=\columnwidth]{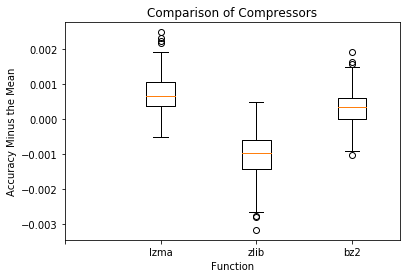}
        \caption{Evaluation of different compressors.}
        \label{figure:compressor}
    \end{minipage}
\end{figure*}

Before proceeding, we would like to stress that neither the cost function nor the compressor are hyperparameters. 
We demonstrate this by showing that the choice of different cost functions and compressors leads to substantially similar results.
In Figure~\ref{figure:metrics}, we apply our algorithm to the discussed  partially overlapping isotropic Gaussian blobs, and evaluate different alternatives for the definition of the cost function $\mathcal{N}(\mathcal{X}, M)$ in~\eqref{eq:n}: arithmetic mean $(\mathcal{I}(\mathcal{X}, M) + \mathcal{S}(\mathcal{X}, M)) / 2$, geometric mean $(\mathcal{I}(\mathcal{X}, M) \times \mathcal{S}(\mathcal{X}, M) )^{1/2}$, harmonic mean $2 / (\mathcal{I}(\mathcal{X}, M)^{-1} + \mathcal{S}(\mathcal{X}, M))^{-1})$, Euclidean distance $(\mathcal{I}(\mathcal{X}, M)^2 + \mathcal{S}(\mathcal{X}, M)^2 )^{1/2}$, sum $\mathcal{I}(\mathcal{X}, M) + \mathcal{S}(\mathcal{X}, M)$, and product $\mathcal{I}(\mathcal{X}, M) \times \mathcal{S}(\mathcal{X}, M)$. The figure shows the extremely limited difference between the different functions (for an easier interpretability of the results, the mean has been subtracted from the values). 
Similarly, Figure~\ref{figure:compressor} shows the performance of our algorithm when using the  LZMA, zlib, and bz2 compressors. We observe that all of them yield similar performance. 
From the above results, we conclude that the performance our algorithm is independent of the specific choice made for either implementation aspect.

In the following, we employ the bz2 as the compressor to approximate the computation of the Kolmogorov complexity, and the harmonic mean as the cost function. The latter has the additional advantage of comparing relative values instead of absolute ones, and therefore makes it possible to get rid of additive constants related to the choice of a specific template for the representation of the tree, as discussed in Section~\ref{sec:practical_impl}.

\begin{figure*}[t]
    \centering
    \includegraphics[width=1.75\columnwidth]{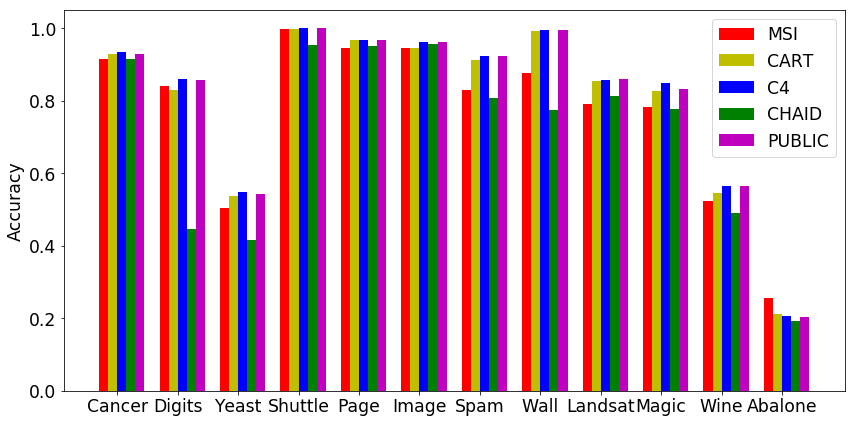}
    \caption{Accuracy over example datasets.}
    \label{figure:real_accuracy}
\end{figure*}

\begin{figure*}[t]
    \centering
    \includegraphics[width=1.75\columnwidth]{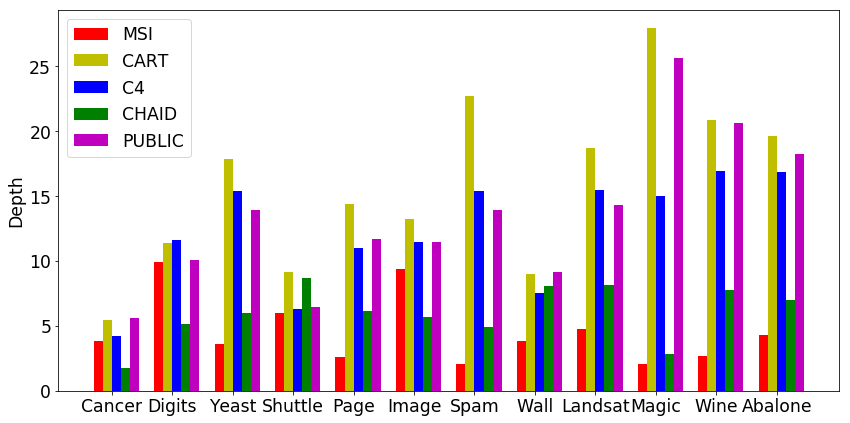}
    \caption{Depth of models over example datasets.}
    \label{figure:real_nodes}
\end{figure*}

\begin{figure*}[t!]
\centering
    \includegraphics[width=1.75\columnwidth]{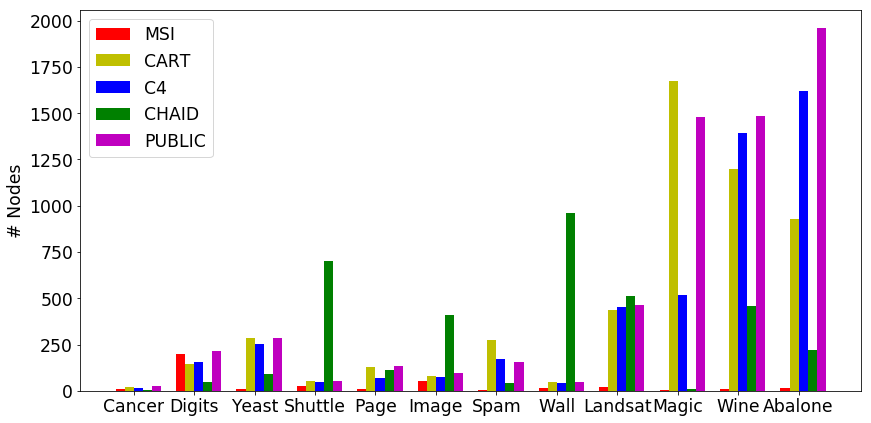}
    \caption{Number of nodes over example datasets.}
\label{figure:real_depth}
\end{figure*}



Finally, we have compared the performance of our algorithm with the performance of four popular decision trees algorithms, CART, C4.5, CHAID, and PUBLIC, with a collection of real datasets. More specifically, we have selected $12$ well known datasets from the UCI Machine Learning Repository~\cite{Dua:2017}, among those with the largest amount of data. The selected ones are: diagnosis of breast cancer (\texttt{cancer}), optical recognition of handwritten digits (\texttt{digits}), predicting protein localization sites in gram-negative bacteria (\texttt{yeast}), classification of NASA space shuttle data (\texttt{shuttle}), classification of blocks in web pages (\texttt{page}), segmentation of outdoor images (\texttt{image}), predicting the age of abalones from physical measurements (\texttt{abalone}), predicting the quality of red and white variants of Portuguese wine (\texttt{wine})~\cite{cortez2009modeling}, filter spam emails (\texttt{spam}), wall-following robot navigation (\texttt{wall}), classification of land use based on Landsat satellite images (\texttt{landsat}), and distinguishing signals from background noise in the MAGIC gamma telescope images (\texttt{magic}). For each dataset, we have repeated the experiment 100 times\footnote{In order to limit the computational complexity, the number of repeated experiments was limited to less than 100 for the CHAID algorithm when applied to some particularly heavy datasets.}, by randomly selecting the training (70\%) and testing (30\%) subsets at each iteration.

In Figure~\ref{figure:real_accuracy} we compare the accuracy of the resulting models obtained by applying CART, C4.5, CHAID, as well as our own new algorithm to the above real datasets. In 3 of the 12 datasets, our algorithm provides better accuracy than CART. In the remaining $9$ cases, the accuracy is fully comparable, and less than $2\%$ smaller than the best of the other algorithms, on average. In Figure~\ref{figure:real_depth} we provide a comparison of the depth of the resulting models. Our algorithm yields a shallower tree than all other algorithms, except for datasets \texttt{Cancer}, \texttt{Digits}, and \texttt{Image}, confirming the non-overfitting properties of our tree construction process.
In Figure~\ref{figure:real_nodes}, we show a comparison of the number of nodes of the resulting models. Only for one of the datasets (\texttt{Digits}), does our model produce a slightly larger tree that those generated by CART, C4.5, CHAID and PUBLIC. In most of the cases, the trees generated by our algorithm have up to three orders of magnitude fewer nodes. Finally, 

Finally, we have compared our algorithm with a post-processing pruning algorithm applied to the tree obtained by CART. In particular, we have applied a cost-complexity pruning guided by a cross-validation to the \texttt{Shuttle} dataset, as implemented by the rpart library in R. The optimal value of the cost-complexity metric is achieved for a tree of 59 nodes, whereas our algorithm, without requiring the optimization of any hyperparameters, obtains a tree of 28 nodes. Both trees achieve the same accuracy.



%
%
\section{Lessons Learned}

The algorithm proposed in this paper has been designed to find a compact model that describes well a dataset (high accuracy) without over-fitting the data. 
%
The experiment described in Figure~\ref{figure:data_error_cart} suggest that our algorithm has lower sensitivity than CART to the errors found in datasets, at least if the number of errors is small compared with the number of valid points. The experiment of Figure~\ref{figure:blobs} shows that the algorithm tends to produce simpler models when the classes that compose the dataset are not linearly separable. These two situations, errors and non-linearity, are common causes of model overfitting when using decision trees. In general, the CART algorithm produces much more complex models than our algorithm in those situations, even when configured to avoid overfitting as much as possible.

As we can see in Figures \ref{figure:real_accuracy}, \ref{figure:real_nodes} and \ref{figure:real_depth}, our new algorithm produces trees with a significantly smaller number of nodes (and depth) than those produced with standard algorithms used in practice, namely CART, C4, CHAID and PUBLIC, without significantly decreasing the accuracy. The experiments have been performed with a collection of datasets resulting from real experiments in order to test the applicability of the new algorithm to practical problems. Given the size of the trees produced, our algorithm is the ideal method to apply in those cases where the interpretability of the results is critical, or where there is a large risk of model overfitting.
 
%
%
\section{Conclusions and Future Work}
\label{sec:conclusionAndFutureWork}

In this paper, we have proposed a new algorithm to build decision trees based on the compressibility of candidate models and the compressibility of the errors generated by those models. The main advantage of the new algorithm is that it does not overfit the training data by design, and additionally it does not require external or ad-hoc procedures to control the overoptimization of the produced model. Moreover, our algorithm does not require the optimization of any hyperparameters. Experimental validation with synthetic and real datasets demonstrates that the accuracy of the new algorithm is similar to the accuracy of well-known decision tree algorithms, like CART, C4.5, CHAID and PUBLIC. Our results show that the proposed algorithm produces models with a considerably smaller number of nodes, without any substantial accuracy decrease.

Future work includes extending the cost function to include a goodness measure for the attributes used at the nodes of the trees, e.g., based on how correlated the values of these attributes are with the target values. 
The early stopping properties of our algorithm can also be applied to other machine learning techniques, e.g., to find optimal deep neural network architectures.


\bibliographystyle{IEEEtran}
\bibliography{IEEEabrv,biblio}

\EOD

\end{document}